\documentclass[sigconf]{acmart}

\usepackage{gensymb}

\usepackage{booktabs} 

\usepackage{times}
\usepackage{xcolor}
\usepackage{soul}
\usepackage{graphicx}
\usepackage{url}
\usepackage{multirow}

\usepackage{balance}

\newcommand{\etal}{\textit{et al}.}
\newcommand{\ie}{\textit{i}.\textit{e}. }
\newcommand{\eg}{\textit{e}.\textit{g}. }

\newcommand*{\origrightarrow}{}

\renewcommand*{\textrightarrow}{\fontfamily{cmr}\selectfont\origrightarrow}

\acmISBN{978-1-4503-6175-0}

\acmConference[IVSP 2019]{International Conference on Image, Video and Signal Processing}{Feb 2019}{Shanghai, China}
\acmYear{2019}
\copyrightyear{2019}

\begin{document}
\title{Learning Autonomous Exploration and Mapping with Semantic Vision}

\author{Xiangyang Zhi}
\affiliation{%
  \institution{ShanghaiTech University}
  \streetaddress{393 Middle Huaxia Road}
  \city{Shanghai}
  \state{China}
  \postcode{201210}
}
\email{zhixy@shanghaitech.edu.cn}

\author{Xuming He}
\affiliation{%
  \institution{ShanghaiTech University}
  \streetaddress{393 Middle Huaxia Road}
  \city{Shanghai}
  \state{China}
  \postcode{201210}
}
\email{hexm@shanghaitech.edu.cn}

\author{S{\"o}ren Schwertfeger}
\orcid{0000-0003-2879-1636}
\affiliation{%
  \institution{ShanghaiTech University}
  \streetaddress{393 Middle Huaxia Road}
  \city{Shanghai}
  \state{China}
  \postcode{201210}
}
\email{soerensch@shanghaitech.edu.cn}

\begin{abstract}
We address the problem of autonomous exploration and mapping for a mobile robot using visual inputs. Exploration and mapping is a well-known and key problem in robotics, the goal of which is to enable a robot to explore a new environment autonomously and create a map for future usage. Different to classical methods, we propose a learning-based approach this work based on semantic interpretation of visual scenes. Our method is based on a deep network consisting of three modules: semantic segmentation network, mapping using camera geometry and exploration action network. All modules are differentiable, so the whole pipeline is trained end-to-end based on actor-critic framework. Our network makes action decision step by step and generates the free space map simultaneously. To our best knowledge, this is the first algorithm that formulate exploration and mapping into learning framework. We validate our approach in simulated real world environments and demonstrate performance gains over competitive baseline approaches.
\end{abstract}

%
%



\maketitle

\section{Introduction}
\label{sec:intro}


For a human who comes to a new environment, one important thing is to look around, acquainting himself with the environment and generating a ``map" in his mind. Similarly, for a mobile robot, 
the process of exploration and mapping, known as autonomous exploration, has been a key problem in robotics for the past several decades. 
In general, the performance of autonomous exploration is measured by the information gain rate, say the size of the explored area, in a unit time. So as to maximize the efficiency of exploration, the robot should try its best to move towards the unexplored space. 

Most of previous work have been focused on the problem setting for robots equipped with active range sensors, such as frontier-based technique~\cite{Yamauchifrontierbasedapproachautonomous1997}. Despite their success, relying on active sensors leads to several limitations, such as high power consumption and restrictive condition for deployment. By contrast, vision-based navigation using cameras has attracted increasing attention due to their low resource footprint and capacity of capturing richer information on the environment~\cite{ohya1998vision}.  Early attempts in this direction employed stereo cameras to obtain 3D information~\cite{sim_autonomous_2006,FraundorferVisionbasedautonomousmapping2012} and to guide the navigation. However, stereo camera systems based on multi-view geometry are less reliable in real-world environments, especially for textureless or reflective surfaces.
Moreover, such geometry-based approaches fail to exploit the semantic cues of the visual scenes, such as object category and spatial layout, which are critical for generating efficient exploration strategies. For instance, if the robot recognizes its current location as a corridor, it may take a specific route to navigate through it. 
  

In this work, we propose to utilize semantic interpretation of the visual scenes along with the camera geometry in tackling the problem of autonomous exploration and mapping. In particular, we aim to exploit the dense semantic segmentation of input frames to build a flexible and confidence-aware map of the traversable space, which in turn yields a rich representation for learning an efficient exploration strategy. Our goal is to maximize the efficiency of autonomous mapping by learning the entire pipeline from an annotated visual environment. 

To this end, we develop a deep neural network that consists of three main modules for vision-based exploration and mapping: visual sensing, map construction and exploring action prediction. Specifically, for each input frame, our first module generates a dense pixel-wise semantic segmentation based on a Fully Convolutional Network with dilation convolution~\cite{chen_deeplab:_2016}. The confidence map of the ground class is sent to the second module, which builds a two-level map representation, ego-centric and bird-view, based on camera pose and scene geometry. The ego-centric map describes the local layout of traversable space while the bird-view representation encodes the global map of the environment. The third module is a convolutional neural network that takes the map representation as input to predict the next movement in robot navigation. A key characteristic of our exploration and mapping network is all three modules are differentiable w.r.t their inputs and parameters, which enable us to train the entire network in an end-to-end fashion.   


We formulate the task of autonomous navigation as a sequential decision process, in which we aim to train the deep neural network to generate an efficient exploration policy. To achieve this, we define a reward function that computes the area that was mapped in a fixed number of movements of the mobile robot, and adopt an actor-critic based policy gradient method to maximize the expected accumulated reward plus a regularization term from the segmentation loss. We evaluate our framework on a large-scale indoor environment~\cite{armeni_cvpr16} and our semantic-driven deep network outperforms other vision-based baselines by a large margin.   

The main contributions of our work are three-fold:
\begin{itemize}
	\item We propose a novel scene representation for vision-based autonomous exploration and mapping, which integrates the semantic image labeling with geometry-based scene models.    
	\item We develop a multi-scale, differentiable map representation that enables us to embed the map construction into a deep neural network. 
	\item We formulate autonomous exploration as a reinforcement learning problem and use a policy gradient method to learn the entire system in an end-to-end manner, which produces a highly efficient exploration policy. 
\end{itemize}

\section{Related Work}
\label{sec:related_work}


There is a large body of literature on autonomous exploration and mapping~\cite{wettach_dynamic_2010,krishnan_visual_2010,SantosSpatiotemporalexplorationstrategies2017}. Typically, the existing pipelines consist of three stages: identifying the unexplored regions, determining which region to go next and navigating to that region. During the exploration, these three stages are executed repeatedly until the map is built. As a partial map is generated during exploration, there are several map-based navigation methods available for the third stage. Hence most prior work focus on the first two stages. 

\noindent\textbf{Range-sensor based exploration:} 
For identifying the unexplored regions, one of the most successful methods is based on frontiers, which are the regions on the boundary between free space and unexplored space. In ~\cite{Yamauchifrontierbasedapproachautonomous1997}, the author assumed that the robot is equipped with range sensors, for instance, sonar or laser range finder, with which the frontiers can be easily identified. Several recent methods employed hand-craft features~\cite{wullschleger_flexible_1999,newman_autonomous_2003,murphy2008using}, which also need range sensors. However, if we rely on cameras only rather than range sensors, the exploration becomes much more difficult due to the loss of 3D information. 

\noindent\textbf{Vision based exploration:}
Sim and Little ~\cite{sim_autonomous_2006} adopted a stereo camera to construct an occupancy map and employ a frontier exploration technique. Santosh \etal ~\cite{santosh_autonomous_2008} segmented floors from the RGB images: first over-segment the image into lots of super-pixels, then, by assuming that a small region directly in front of the robot is free space, find similar super-pixels and label them as floor. With the segmented floor, the robot can move without collision. However, one obvious disadvantage is that the floor segmentation method is not robust, especially when the color of the floor varies widely. Fraundorfer \etal ~\cite{FraundorferVisionbasedautonomousmapping2012} conducted visual SLAM with a stereo camera mounted on a MAV, generating a 3D occupancy map, and then by reducing the 3D occupancy map to 2D occupancy map, used frontier-based exploration method proposed in ~\cite{Yamauchifrontierbasedapproachautonomous1997}. 

\noindent\textbf{Exploration strategy:}
Given the interesting regions of the environment, the next step for a mobile robot is to select one region to go. A naive but efficient algorithm is to choose the nearest region, which is adopted widely in literature~\cite{Yamauchifrontierbasedapproachautonomous1997,wullschleger_flexible_1999,murphy2008using,santosh_autonomous_2008,FraundorferVisionbasedautonomousmapping2012}. There are also several approaches~\cite{oswald_speeding-up_2016,gonzalez-banos_navigation_2002} that consider the utility, i.e., information gain when the robot reach one region of interest, in determining the exploration strategy. However, these existing approaches divide the exploration into several steps, while in this paper we propose an end-to-end exploration pipeline.

\noindent\textbf{Reinforcement learning:}
Reinforcement Learning (RL) has been 
successfully applied to a variety of robot applications. 
\cite{mahadevan_automatic_1992} proposed a RL-based method to make a robot push boxes autonomously. \cite{kim_autonomous_2004} utilized RL to control a helicopter. In \cite{michels_high_2005}, Michels \etal~used RL to detect obstacles with a monocular camera. RL has also been successful in locomotion and motor control for various type of robots~\cite{kohl_policy_2004,noauthor_reinforcement_2008}. 

Recently, due to the unprecedented success of deep learning, deep RL has shown remarkable development in tackling complicated problems. 
\cite{mnih_human-level_2015} proposed Deep q-learning (DQN) which achieves higher scores than human players in ATARI games. Furthermore, \cite{silver_mastering_2016} proposed a deep RL algorithm with Monte-Carlo tree search (MCTS), which defeats the world champion in the game of Go. To improve learning efficiency, four asynchronous gradient descent methods for deep RL are discussed in \cite{mnih_asynchronous_2016}, including the Asynchronous Advantage Actor-Critic (A3C), which is widely adopted in solving various RL problems. Moreover, Deep RL has also been used in robotics with end-to-end training, such as robotic manipulation~\cite{levine_end--end_2015,levine_learning_2016} and robot navigation~\cite{zhu_target-driven_2016,mirowski_learning_2016}.

\noindent\textbf{Relationship to contemporary work:}
To our best knowledge, there is little work in learning-based exploration and mapping so far. On the other hand, Deep RL has been used in robot navigation, which is a related problem as it also needs path planning~\cite{gupta_cognitive_2017,gupta_unifying_2017}. 
Among them, perhaps the most related work is~\cite{gupta_cognitive_2017}, in which a mapper also predicts the free space of the environment and accumulates the free space to a map representation. However, our work differs in the following two aspects: first we build an explicit map on traversable space using the semantic segmentation to predict free space and then transforming the free space to bird-view grid map by perspective projection; second, instead of using imitation learning~\cite{ross_reduction_2010}, which is difficult to apply to our problem, we adopt an efficient RL approach, A3C, to learn the full pipeline. 
  

\section{Problem Setup}
\label{sec:problem_setup}

\begin{figure*}
	\centering
	\includegraphics[width=1.\linewidth]{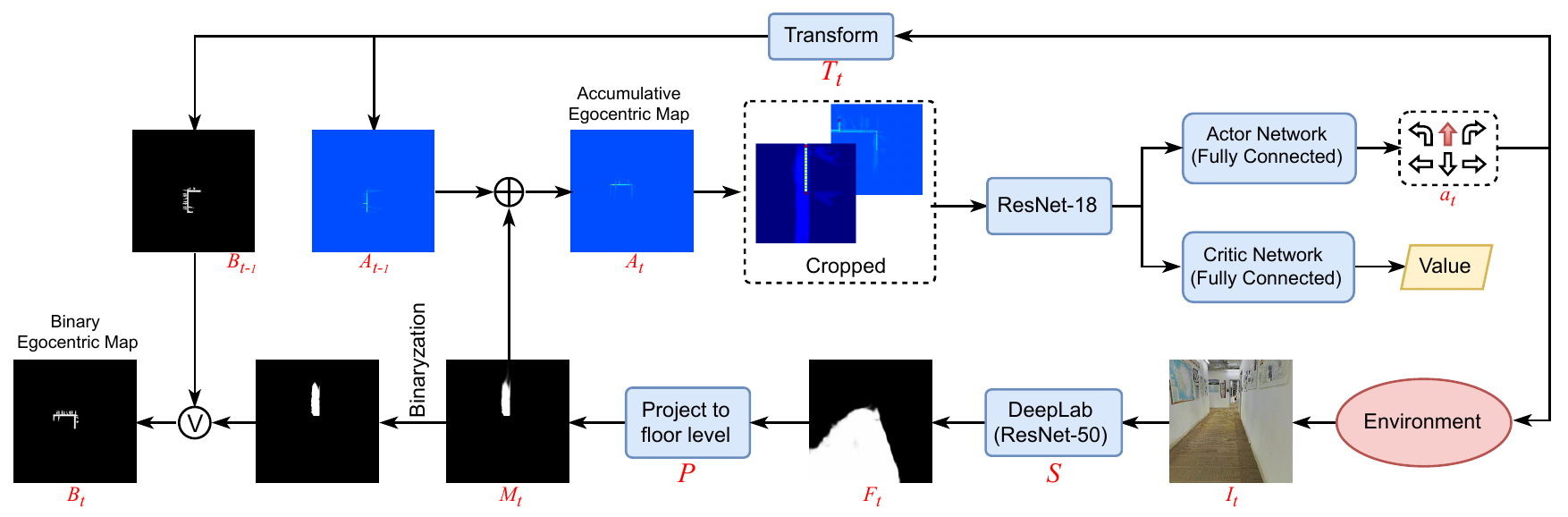}
	\caption{Diagram of proposed exploration and mapping method. }
	\label{fig:diagram}
\end{figure*}

As it is very difficult to train on a real robot, we adopt a simulation-based strategy in this work.  
A number of indoor stimulation environments have been developed, such as AI2-THOR~\cite{ai2thor} and SUNCG ~\cite{song2016ssc}. However, the scenarios of these datasets are synthetic, thus showing significant differences to reality. Inspired by the work of Gupta \etal~\cite{gupta_cognitive_2017}, we run our algorithm on the Stanford large-scale 3D Indoor Spaces (S3DIS) dataset~\cite{armeni_cvpr16}. In contrast to AI2-THOR and SUNCG, this dataset is collected by scanning the real environment with a Matterport scanner\footnote{\url{https://matterport.com/}}. It consists of 6 large-scale indoor areas that originate from 3 different buildings of mainly educational and office use. For each area, there is a 3D reconstruction with textured mesh, as well as the corresponding 3D semantic mesh. 

We simulate a robot in that environment. Identical to the work of Gupta \etal, we simplify the robot as a cylinder, equipped with a RGB camera mounted on top of the robot rigidly. In addition, to focus on the high-level action decision, we also assume that the robot has accurate motion and perfect odometry, for example, by employing visual odometry. We assume that the robot is omni-directional using Mecanum wheels or omni wheels, so we specify six actions for the robot: move forward $x$ cm, move backward $x$ cm, move left $x$ cm, move right $x$ cm, turn left by angle $\theta$ and turn right by angle $\theta$. 
In this paper, $x$ and $\theta$ are set to 40 and $\pi/2$, respectively. 
With the above setting, we can actually discretize the traversable space into a directed graph of which the nodes denote the traversable spots and directed edges indicate the robot can move from one node to another by one of the above six actions.

\section{Methodology}
\label{sec:methodology}


To learn an efficient autonomous exploration and mapping strategy, we develop a deep neural network that consists of three modules according to their functions, and can be trained end-to-end.  
Figure \ref{fig:diagram} illustrates the overview of our network architecture. 
The first module of our pipeline is a semantic segmentation network, which is designed to obtain semantic features from the RGB image, especially the free space. The second module is mapping of free space, which extends a bird-view map step by step. And the third module is an action decision network determining the next movement of the robot. After executing the action, the robot moves to a new position, starting the whole procedure again. The entire network is learned based on the actor-critic framework. We now introduce the three modules one by one.
\subsection{Semantic Segmentation}
As the robot needs to build up a map of the environment, it is necessary to extract free space from the image captured by the camera. To achieve this, we first employ a semantic segmentation network to extract pixel-level semantic information from the image. 
In addition to free space (represented by the floor class), we also generate segmentation of other object classes as they provide informative context cues. 
Specifically, we group the rest of object labels into three super-classes and segment each image into four semantic categories. As we have the ground-truth semantic labeling in the training dataset, we compute the loss of the semantic segmentation network and integrate it into the exploration reward.


Mathematically, at time step \textit{t}, the robot takes a photo of the environment, say $I_t$. After the semantic segmentation function $S$, we get free space image $F_t$, \ie $F_t=S(I_t)$.

\subsection{Mapping}
\label{sec:mapping}
For an indoor scenario, 2D grid map is usually enough for most applications. In this paper, the map generated is also in 2D. Although the free space has been segmented in the image, we cannot map it directly, because the depth of the pixels are unknown. However, as the robot operates indoor, it is reasonable to assume that the ground level is planar. With this assumption, we can determine the floor corresponding to the free spaces in the images. 

\begin{figure}
	\centering
	\includegraphics[width=1.\linewidth]{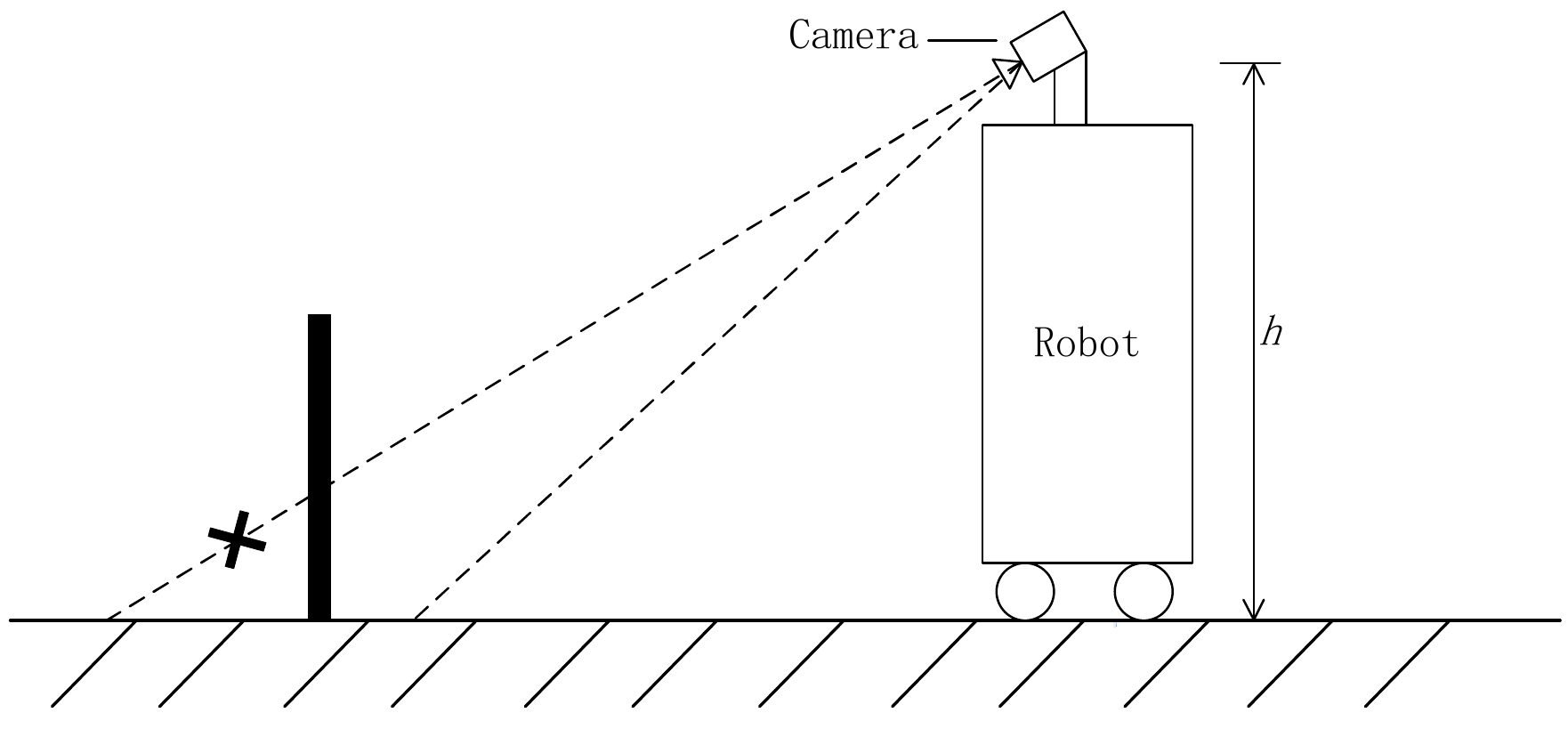}
	\caption{Sketch of free space projection.}
	\label{fig:projection}
\end{figure}

We assume the camera is well calibrated, \ie the intrinsic and extrinsic parameters, here the transformation between camera frame and robot frame, are known. Moreover, as we also assume the odometry is perfect, which means that the transform between world frame and camera frame is known. As shown in Figure~\ref{fig:projection}, since the height and orientation of the camera is known and fixed, we can determine the corresponding free cells of the 2D grid map. We denote this projection function as $P$ and the bird-view map as $M_t$ at time step $t$, \ie $M_t = P(F_t)$. It is noteworthy that this projection only fits the free space, \ie floor, because any object which is higher than floor will be projected to the wrong location in this way, \eg the black obstacle shown in Figure~\ref{fig:projection}. This is the reason why we only project free space rather than all object classes.

We maintain two egocentric maps in this module. One is binary, \ie the value of a cell of the map is 1 if it's free or 0 if not. This binary egocentric map (BEM) is the explicit output of our method. On the other hand, we also have an accumulative egocentric map (AEM) whose cells are accumulating when a new projection hits them. The AEM is generated as the input for the action decision network. We do not normalize the AEM, as it implicitly encodes the visited times of areas, which could be very useful for the robot to decide its next actions. In addition, we note that the robot's trajectory also includes important information: the places that have been visited. Hence, when a robot arrives at a new position, we increase the cells of the AEM which the robot stands on by a large value (20.0 in this paper). Besides indicating the trajectory of the robot, this operation also enhances the map of free space, since obviously the places a robot passed though are traversable. On the other hand, a place that the robot tried to reach but failed to do so is occupied, and thus we subtract the corresponding cells of the AEM by a large value (here it is also 20.0). This operation helps avoiding collision to the obstacles at the beginning of the training.

We denote AEM and BEM by $A_t$ and $B_t$ at time step $t$. At time step $t$, if the robot takes action $a_t$ according to $A_{t-1}$, AEM and BEM should be updated by a transform $T_t$, which can be calculated from $a_t$. Then $A_t$ can be updated as
\begin{equation}
A_t = T_t(A_{t-1}) + M_t.
\end{equation}
As BEM is a binary map, we apply a threshold to binarize $M_t$. To reduce false positives, which have large impact on the map quality due to incorrect projections, we set the threshold to 0.8. So $B_t$ is updated as
\begin{equation}
	B_t = T(B_{t-1}) \vee (M_t>0.8)
\end{equation}

The size of AEM and BEM, theoretically, should be as large as possible to ensure the robot stays in the map and to avoid information loss. However, due to the constraints on computation resources in practice and the physical size of the area to be explored, we set a maximum size of AEM and BEM according to those factors. 

\subsection{Action Decision}
To generate the full map, a critical step is to make action decision for navigation. A good navigation strategy should always guide the robot to unexplored areas and avoid collisions on the way. To achieve this, we adopt a neural network with fully connected layers to predict the next movement based on the map inputs.
As this is a typical sequence decision problem, we utilize an actor-critic framework to learn the action decision network and to fine-tune the rest of the pipeline. 
We will focus on two key elements of the reinforcement learning, the reward and state design, in this section. 

\noindent\textbf{Reward Design.} The goal of the exploration is to explore an unknown environment as quickly as possible. As such, we design a reward function with the following three components.   
The first component of our reward is the area of free space that the robot identifies given the time spent. This reward allows us to guide RL training at each step as it can be computed densely during the exploration. Given a fixed time budget, it also enables us to parallelize the training of several robots in a distributed manner due to the fixed length of each episode. 
On the 2D grid map, the reward can be easily computed as the additional number of free grids of the map. 
In addition, we punish the false positive of free space segmentation by giving a negative reward in that case. This negative reward is quantified by the number of pixels of the input image which are labeled as free incorrectly. Furthermore, to make the robot learn to avoid collisions, which obviously have a negative impact for exploration, a small negative reward will be given if a collision happens. 

\noindent\textbf{State Design.} As mentioned before, AEM is the input of the action decision module. 
%
However, AEM is very large, most information is useless for the current action decision, so as to utilize the AEM efficiently, we crop and scale the map around the center. The cropped AEM with the steps of 1 and 3, respectively, have the same size: $257\times 257$. Then we concatenate these two AEMs, so finally the size is $257\times 257\times 2$.
The AEM is simple but very powerful, because it not only functions as a free space map, but also contains information about the trajectory of the robot and occupied space. Moreover, we tried adding image features extracted by the segmentation network along with features extracted from AEM, but did not find any performance improvement. This indicates that the AEM is sufficient as a state in this task.

\section{Experimental Results}
\label{sec:experiments}


\noindent\textbf{Implementation Details:} The semantic segmentation network is based on DeepLab~\cite{chen_deeplab:_2016} using the ResNet-50~\cite{he2016deep}. We do not employ any label refinement module such as Conditional Random Field. The action decision network is a ResNet-18.

The RGB image size is $257\times 257\times 3$. Unlike many deep RL approaches, we only utilize the current frame as the input rather than stack several historical frames. One grid of AEM or BEM represents a $5cm\times 5cm$ square on the ground floor.

Our model is trained asynchronously with 4 parallel GPU workers using TensorFlow~\cite{abadi2016tensorflow}. We use A3C~\cite{mnih_asynchronous_2016} as the training protocol of the actor-critic network, and train the full model in an end-to-end fashion. We use ADAM~\cite{kingma_adam:_2014} to optimize our loss function with an initial learning rate of 2e-5. The loss function comprises the policy loss of the actor and value loss of the critic in the actor-critic network, regularized by the cross entropy loss of the semantic segmentation. We balance them with scale factors. The episode length is fixed to 200 in the experiments, and the number of local steps is 20, \ie a single reward directly affects the values of 20 preceding state action pairs. We set the discount factor of rewards to 0.99.

There are 6 areas in the S3DIS dataset, but the floors of \textit{area2} is not in one level, so we exclude it and conduct experiments on the other 5 areas. \textit{area1}, \textit{area3}, \textit{area4} and \textit{area6} are used as training dataset, and \textit{area5} is used as evaluation dataset. During training, we run several simulated robots on one GPU worker in parallel, and distribute them evenly over the four areas. We believe this kind of training reduces the bias of a batch of data which may impede gradient descend.

\subsection{Comparison to Baseline}
To our best knowledge, there are no similar learning-based exploration methods like ours, so as to evaluate our approach, we implement two kinds of baselines using RGB image and depth image as input, respectively. Both of two baselines simply use the current frame as the state of the actor-critic network, inside which is ResNet-50 followed by LSTM. With regard to free space segmentation, it is very easy for the depth image, but difficult for RGB image. So as to make the RGB-based baseline more competitive, we utilize the ground truth as free space segmentation result for it. 
Similarly, a third baseline method is also using the ground truth segmentation to crate the map but a random action policy.  We called the three baselines ``RGB+LSTM", ``depth+LSTM" and ``Random". Also, the range of available commercial depth cameras is 8m at most (Kinect 2), so as to show the performance of a real depth camera, our fourth baseline is ``depth+LSTM" where the range of depth camera is limited to 8m, named  ``depth+LSTM (8m)".

Obviously, it will be more difficult for the robot to start in a room than in a corridor, because there can be much more stuff inside a room impeding the robot's movement. To compare the performance of different algorithms, we separate the evaluation tasks by the place where the robot starts, which consists of 3037 different starting points in corridors and 2073 in rooms. Each method runs for one episode (200 steps) by starting at all points, and the final score is the mean of that. Regarding evaluation metrics, the score are the free areas that are labeled correctly subtracting the occupancy areas that are labeled as free falsely. The experiment results are listed in Table~\ref{fig:results}. The performance is only worse than that of ``depth+LSTM", which actually cannot be achieved with an available depth sensor. In comparison to ``RGB+LSTM", which uses the same sensor, our method improves the area of explored free space by 49.7\% and 51.3\% when starting in rooms and corridors, respectively.
\begin{table}[t]
	\caption{Average Performance of Different Methods}
	\label{fig:results}
	\centering
	\begin{tabular}{|c|c|c|}
		\hline \multirow{2}{*}{Method} & \multicolumn{2}{c|}{Area of Explored Free Space ($m^2$)}  \\ \cline{2-3}  & Start in rooms & Start in corridors \\
		\hline
		\hline Random Action & 57.34 & 102.80 \\ 
		\hline RGB+LSTM & 102.55 & 118.51 \\
		\hline depth+LSTM (8m) & 137.40 & 161.17 \\ 
		\hline Ours & 153.49 & 179.27 \\
		\hline depth+LSTM & 178.85 & 197.85 \\ 
		\hline 
	\end{tabular} \vspace{-1mm}	
\end{table}

\begin{figure}[!t]
	\centering
	\includegraphics[width=1.0\linewidth]{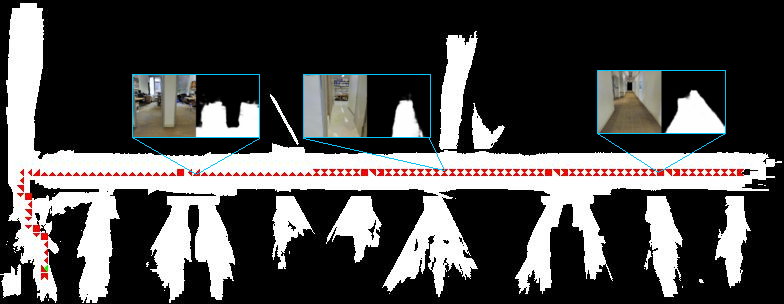}\vspace{0.5mm}\\
	\includegraphics[width=1.0\linewidth]{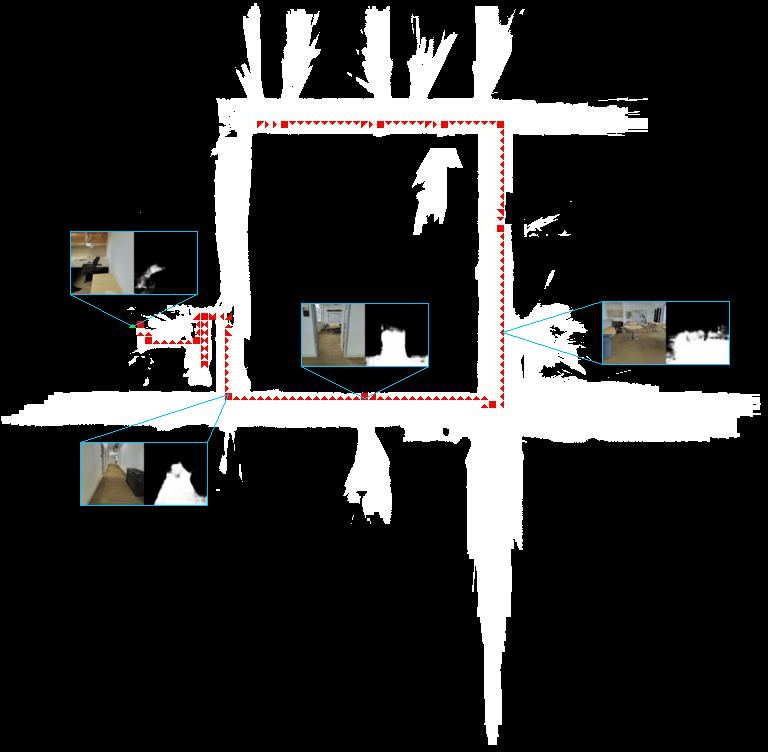}\vspace{0.5mm}\\
	\includegraphics[width=1.0\linewidth]{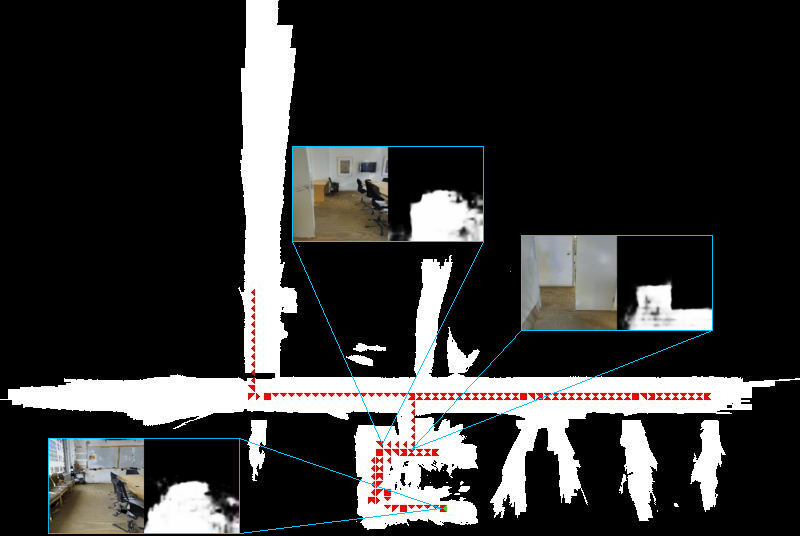}
	\caption{Three representative maps generated by our method. White regions show free space, and red triangles form the trajectories of the robot.  Each triangles denotes one position and orientation of the robot, and the longest edge opens towards the orientation of the robot. The green triangles are starting points. We also embed some obtained first-person images and corresponding free space images segmented by our method. }\vspace{-2mm}
	\label{fig:map_trajectory}
\end{figure}

\begin{figure}[t]
	\centering
	\includegraphics[width=0.18\linewidth]{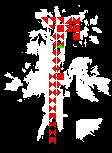}
	\includegraphics[width=0.553\linewidth]{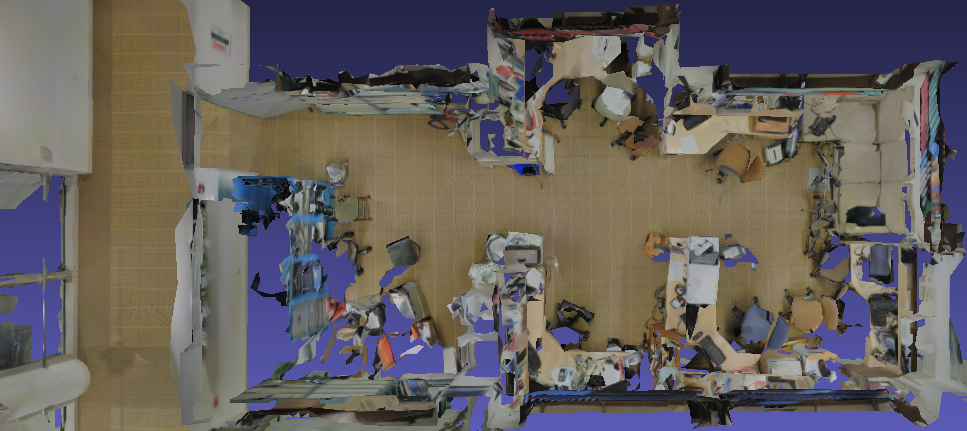}
	
	\includegraphics[width=0.250\linewidth]{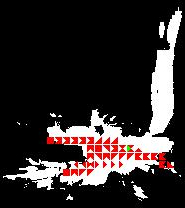}
	\includegraphics[width=0.445\linewidth]{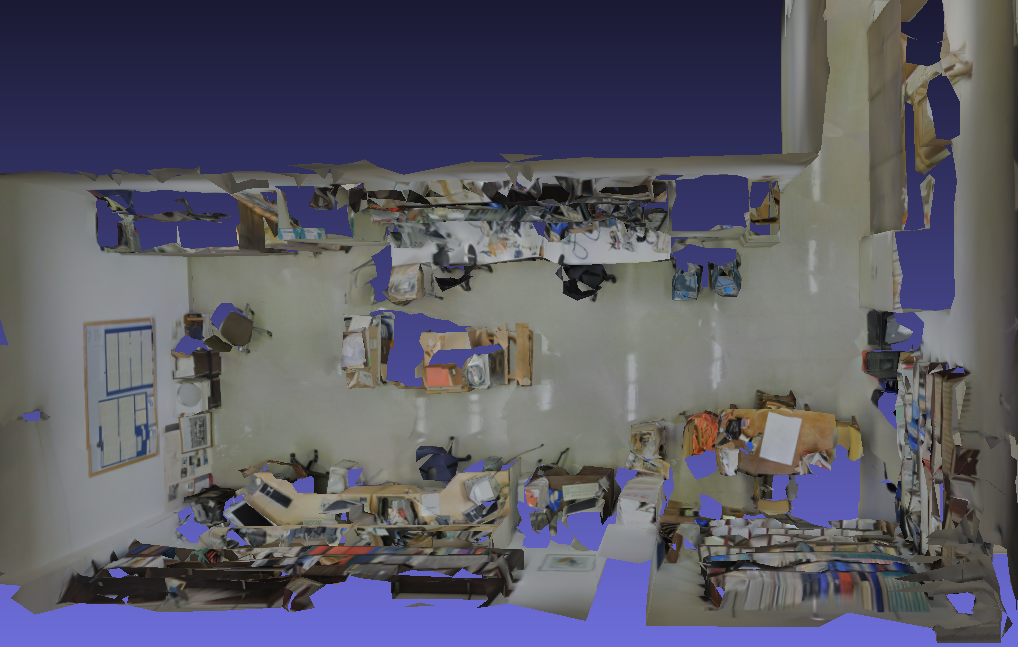}	
	\caption{Two cases in which our algorithm performs bad because it doesn't exit the room. Left figures are the maps and trajectories whose format is described in Figure~\ref{fig:map_trajectory}, and the figures on the right are the bird-view of corresponding scenarios of the maps on the left.}
	\label{fig:bad_cases}
\end{figure}

Intuitively, when the robot starts in rooms, it should get out to explore more space, so finding the way out of rooms is quite an important ability the robot need to have. In Figure~\ref{fig:map_trajectory}, we show three representative maps and the trajectories generated by our method. The figures illustrate that our algorithm empowers the robot to find the way even in deep positions of rooms. Of course, our method is not perfect, there are some bad cases in which the robot gets stuck in rooms, as shown in Figure~\ref{fig:bad_cases}. To analysis the reason resulting in such bad cases, the perhaps one is that the rooms are very messy, occupying by various stuff which makes it more difficult to understand the current scenes and move.

\begin{figure}[t]
	\centering
	\includegraphics[width=1.\linewidth]{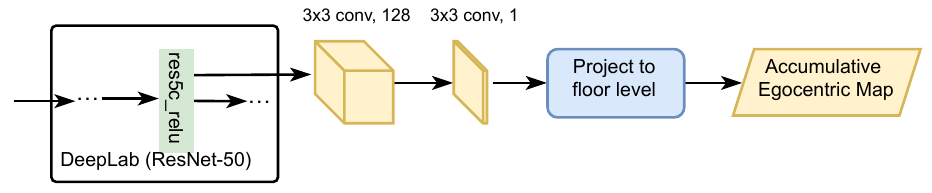}
	
	\caption{The diagram of semantic features projection.}
	\label{fig:projection_features}\vspace{-2mm}	
\end{figure}

\begin{table}[t]
	\caption{Performance of Ablated Versions of our Method}
	\label{fig:ablation}
	\centering
	\begin{tabular}{|c|c|c|}
		\hline \multirow{2}{*}{Method} & \multicolumn{2}{c|}{Area of Explored Free Space ($m^2$)}  \\ \cline{2-3}  & Start in rooms & Start in corridors \\
		\hline
		\hline  Projecting Features & 18.33 & 55.83 \\
		\hline CNN Projection & 84.40 & 102.65 \\
		\hline 
	\end{tabular}
\end{table}

\subsection{Ablation Study}
We also studied ablated versions of our proposed method, the performance of whom is listed in Table~\ref{fig:ablation}. 

\textit{Projecting Features.} There is concern that whether encoding more semantic features can improve the performance. To study that, we replace free space with semantic features, as illustrated in Figure~\ref{fig:projection_features}.

\textit{CNN Projection.} As described in Section~\ref{sec:mapping}, our method utilizes the prior knowledge of the robot setup, which makes map projection quite efficient. We would like to know that whether this projection procedure can be learned using CNN, which will make the calibration of the camera unnecessary.

``Projecting Features" is hard to train, even diverged at the end of training. We try to,  but cannot get comparable performance. By comparison, ``CNN Projection" performs a little better than ``Random Action", but still falls behind the proposed method large.

\subsection{Noisy Localization Experiments}
We assume perfect odometry in our work, but for a real robot operating in real scenarios, perfect odometry is quite hard to obtain. So as to validate the robustness to localization noise, we add random noise to the robot's pose and find out the effect to our method. Obviously, the direct effect of localization noise to our method is the egocentric map. If localization is noisy, the map will be bend, affecting the exploration actions. 

We assume that the noise is with normal distribution, \ie its probability density is
\[
f(|\mu, \sigma^2) = \frac{1}{\sqrt{2\pi \sigma^2}}\exp^{-\frac{(x-\mu)^2}{2\sigma^2}}
\]
where $\mu$ is the mean and $\sigma$ is the standard deviation.

We do not re-train the network with noisy localization, but just add noise to test experiments. Moreover, the noise is added every step, so it will accumulate gradually, and may become quite large at the end of an episode.

Translation noise and orientation noise are both tested, and also different noise levels. We reduce the number of the different starting points, due to time constraints. Thus the resulting performance is slightly different to Table~\ref{fig:results}. In the translation noise experiments, the number is 1270, including points in rooms and corridors. In the orientation noise experiments, the number is 300, also including points in rooms and corridors. The starting points are selected randomly and not changed for different noise levels experiments.

The experiment results are listed in Table~\ref{fig:noisy_localization}. Inevitably, the performance drops after adding noise, however, the decrease is within an acceptable range, and our method with localization noise still outperforms the other baselines. In comparison, the impact to episodes starting in rooms is larger than that starting in corridors. This situation is predictable. When the robot is at a room, it needs a more precise map to help itself avoid collisions and to find the way out of the room. However, in corridors, the free space is wider, and the robot usually follows corridors easily.

These experiments indicate that our method is robust to reasonable localization noise. This result is not unexpected. The free space segmentation is not perfect, so there is always noise in the egocentric map, our network owns the power to extract useful information from noisy map.
\begin{table}[h]
	\newcommand{\tabincell}[2]{\begin{tabular}{@{}#1@{}}#2\end{tabular}}
	\caption{Average Performance of Proposed Method under Different Noise Level. In left column, TN is abbreviation of `Translation Noise', ON is abbreviation of `Orientation Noise'. Then is mean ($\mu$) of noise, followed by standard deviation($\sigma$). }
	\label{fig:noisy_localization}
	\centering
	\begin{tabular}{|c|c|c|}
		\hline \multirow{2}{*}{Noise Level} & \multicolumn{2}{c|}{Area of Explored Free Space ($m^2$)}  \\ \cline{2-3}  & Start in rooms & Start in corridors \\
		\hline
		\hline TN: 0cm, 0cm & 148.88 & 181.90 \\ 
		\hline TN: 0cm, 3cm & 143.97 & 182.44 \\ 
		\hline TN: 0cm, 5cm & 143.95 & 180.34 \\
		\hline TN: 2cm, 3cm & 142.96 & 180.13 \\ 
		\hline TN: 2cm, 5cm & 140.55 & 175.96 \\ 
		\hline \\
		\hline ON: 0.0\degree, 0.0\degree & 146.04 & 176.23 \\
		\hline ON: 0.0\degree, 0.5\degree & 137.36 & 172.46 \\ 
		\hline ON: 0.2\degree, 0.5\degree & 131.32 & 171.20 \\ 
		\hline \tabincell{l}{ON: 0.0\degree, 0.5\degree \\ TN: 0cm, 2cm} & 133.08 & 172.20 \\
		\hline \tabincell{l}{ON: 0.2\degree, 0.5\degree \\ TN: 0cm, 2cm} & 128.97 & 170.25 \\
		\hline 
	\end{tabular} 
\end{table}

\subsection{Rewards Curve}
\begin{figure}
	\centering
	\includegraphics[width=1\linewidth]{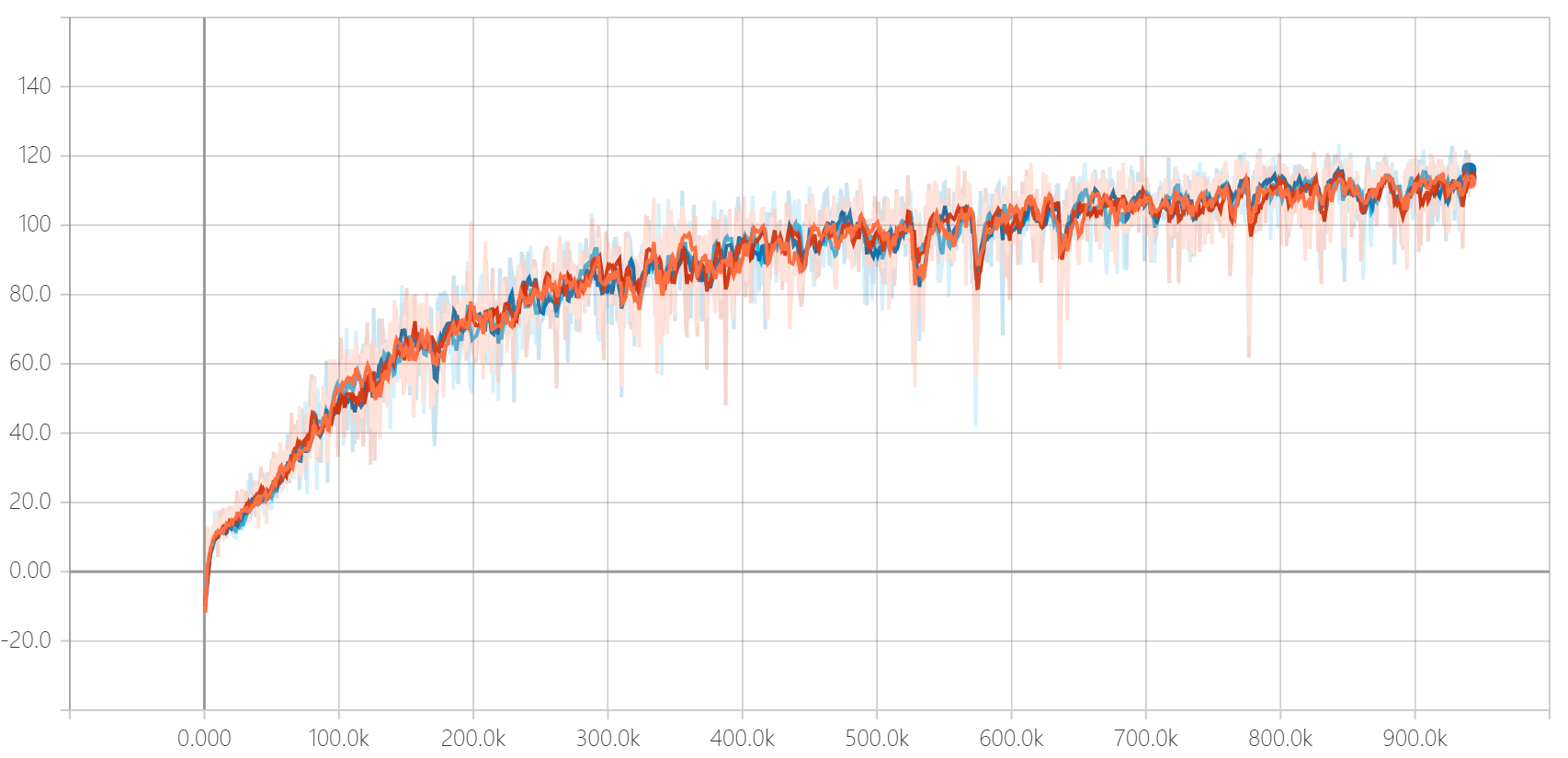}
	\includegraphics[width=1\linewidth]{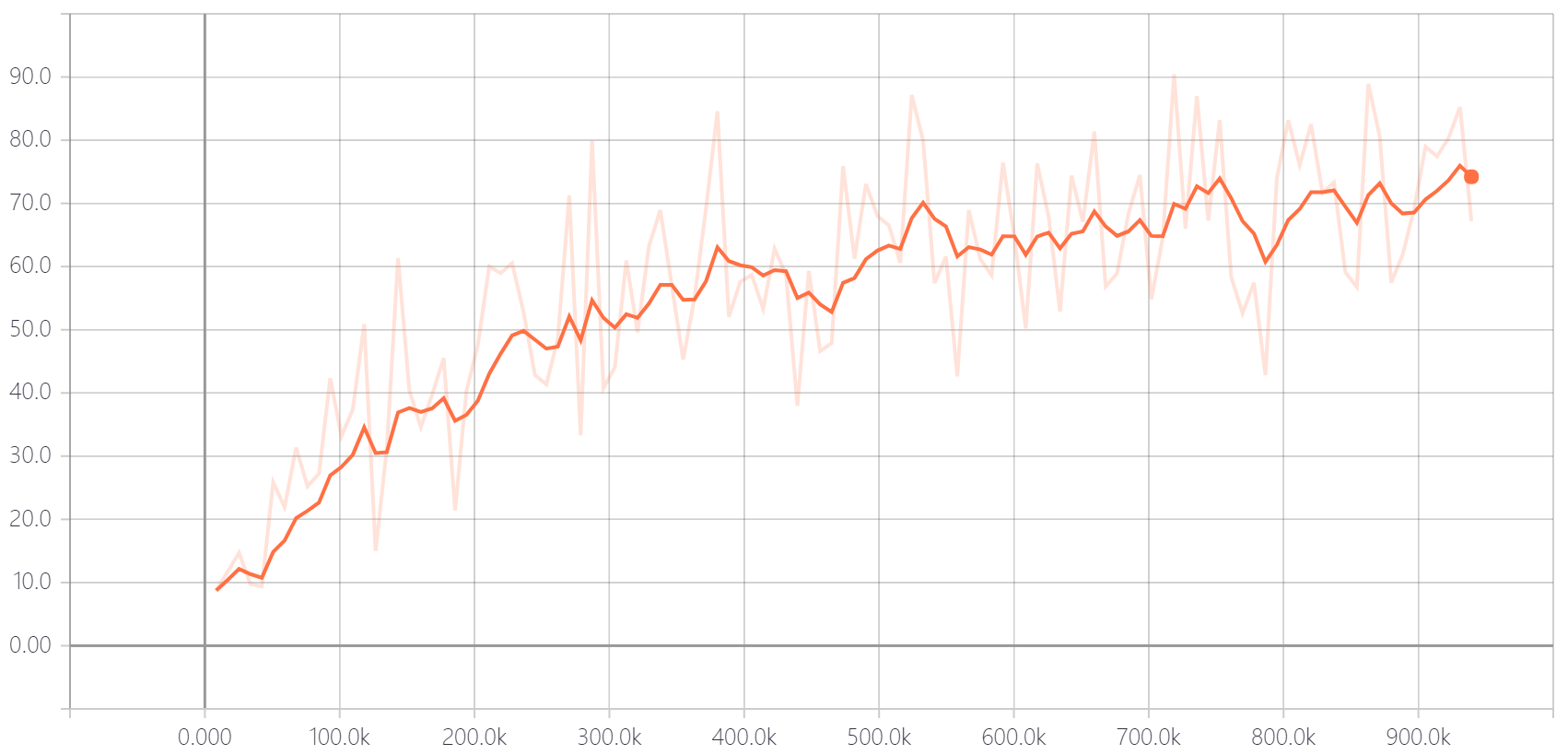}
	\caption{The rewards Curve of training and test. The top figure is for training, which includes four curves, corresponding to four GPU workers. The bottom figure is for test. The x-axis is the overall steps of training and the y-axis is the reward. The transparent curves illustrate original data, while the bold curves are the smoothed version with the factor 0.85.}
	\label{fig:testrewards}
\end{figure}

We show the rewards curve of training, and testing results on validation dataset (area5) while training, see Figure~\ref{fig:testrewards}. Please don't confuse the rewards and area of explored free space in  Table~\ref{fig:noisy_localization}, they are different. As clearly showed, we stop training a little bit early, the rewards are still growing up, so we may obtain slight better performance if we trained longer time.

\section{Conclusion}
\label{sec:conclusion}

This work presents an autonomous exploration and mapping algorithm for indoor scenes using visual inputs only. Our method formulates autonomous exploration and mapping as learning problem. Our network can be trained end-to-end, and we utilize deep reinforcement learning to train an exploration policy with additional guidance from semantic segmentation. Our network can generate exploration actions and the free space map, thus achieve the two core functions of exploration and mapping simultaneously. We conduct our experiments on real world datasets obtained by Matterport, which is very challenging because of various objects crowding in the scenarios. We even perform training and testing on different areas, and experiments show that our method outperforms vision based baseline approaches significantly and is also competitive to depth based methods, which obviously have big advantage with 3D information.

One main limitation of our method is that we assume perfect odometry, which is usually hard to achieve with a real robot in real world. But our explicit free space map makes it easy to cooperate with analytic mapping methods, \eg visual SLAM, which may help solve the problem partially. So we would like to leave it as future work to deploy our algorithm on a real robot.

\bibliographystyle{ACM-Reference-Format}
\bibliography{ijcai18}

\balance

\end{document}